\documentclass[10pt,twocolumn,letterpaper]{article}

\usepackage{cvpr}
\usepackage{times}
\usepackage{epsfig}
\usepackage{graphicx}
\usepackage{amsmath}
\usepackage{amssymb}
\usepackage{wrapfig}
\usepackage{multirow}
\usepackage{subcaption}
\newtheorem{remark}{Remark}

\usepackage[breaklinks=true,bookmarks=false]{hyperref}

\cvprfinalcopy 

\def\cvprPaperID{1582} 

\begin{document}

\title{Scale-Equalizing Pyramid Convolution for Object Detection}

\author{Xinjiang Wang\thanks{equal contribution}, Shilong Zhang$^\ast$, Zhuoran Yu, Litong Feng, Wayne Zhang\\
SenseTime Research\\
{\tt\small \{wangxinjiang, zhangshilong, yuzhuoran, fenglitong, wayne.zhang\}@sensetime.com}
}

\maketitle
\thispagestyle{empty}
\begin{abstract}
Feature pyramid has been an efficient method to extract features at different scales. Development over this method mainly focuses on aggregating contextual information at different levels while seldom touching the inter-level correlation in the feature pyramid. Early computer vision methods extracted scale-invariant features by locating the feature extrema in both spatial and scale dimension. Inspired by this, a convolution across the pyramid level is proposed in this study, which is termed pyramid convolution and is a modified 3-D convolution. 
Stacked pyramid convolutions directly extract 3-D (scale and spatial) features and outperforms other meticulously designed feature fusion modules. 
Based on the viewpoint of 3-D convolution, an integrated batch normalization that collects statistics from the whole feature pyramid is naturally inserted after the pyramid convolution. 
Furthermore, we also show that the naive pyramid convolution, together with the design of RetinaNet head, actually best applies for extracting features from a Gaussian pyramid, whose properties can hardly be satisfied by a feature pyramid. In order to alleviate this discrepancy, we build a scale-equalizing pyramid convolution (SEPC) that aligns the shared pyramid convolution kernel only at high-level feature maps. Being computationally efficient and compatible with the head design of most single-stage object detectors, the SEPC module brings significant performance improvement ($>4$AP increase on MS-COCO2017 dataset) in state-of-the-art one-stage object detectors, and a light version of SEPC also has $\sim3.5$AP gain with only around 7\% inference time increase. The pyramid convolution also functions well as a stand-alone module in two-stage object detectors and is able to improve the performance by $\sim2$AP. The source code can be found at \url{https://github.com/jshilong/SEPC}. 
\end{abstract}
\begin{figure}
\begin{center}
\includegraphics[width=\linewidth]{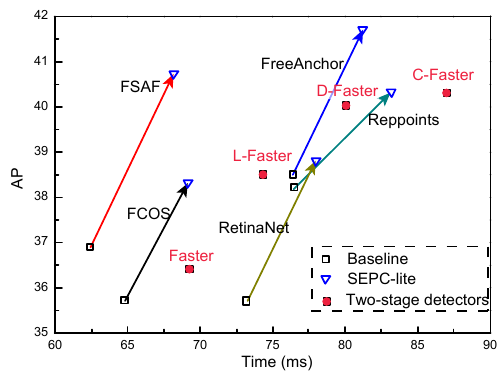}
\end{center}
\caption{\small{Performance on COCO-\texttt{minival} dataset of pyramid convolution in various single-stage detectors including RetinaNet \cite{lin2017focal}, FCOS \cite{tian2019fcos}, FSAF \cite{zhu2019feature}, Reppoints \cite{yang2019reppoints}, FreeAnchor \cite{zhang2019freeanchor}. Reference points of two-stage detectors such as Faster R-CNN (Faster) \cite{ren2015faster}, Libra Faster R-CNN (L-Faster) \cite{pang2019libra}, Cascade Faster R-CNN (C-Faster) \cite{cai2018cascade} and Deformable Faster R-CNN (D-Faster) \cite{zhu2019deformable} are also provided.  All models adopt ResNet-50 backbone and use the \texttt{1x} training strategy.}}
\label{fig:boosts}
\end{figure}
\section{Introduction}
An object may appear in vastly different scales in natural images and yet should be recognized as the same. The scales can easily vary by more than 1 magnitude in natural images \cite{singh2018analysis}, which presents as a challenging task in various computer vision tasks such as object detection.
Extensive research has focused on this issue. Multi-scale training \cite{dai2016r} is a direct solution to scale changes by letting the network memorize the patterns at different scales. Multi-scale inference \cite{ najibi2018autofocus} shares the same idea with traditional image pyramid methods \cite{lowe2004sift, pedersoli2009high}. However, the image pyramid method is time-consuming since multiple inputs are necessary. Intrinsic feature pyramid \cite{liu2016ssd} in CNNs at different stages provides an efficient alternative to image pyramid. Each level of the downsampled convolutional features corresponds to a specific scale in the original image. 
However, there exists a semantic gap between each two levels in feature pyramid. To alleviate the discrepancy, different feature fusion strategies have been proposed, including top-down information flow \cite{lin2017feature, fu2017dssd}, an extra bottom-up information flow path \cite{liu2018path, woo2019gated, kong2017ron}, multiple hourglass structures \cite{newell2016stacked,zhao2019m2det}, concatenating features from different layers \cite{li2017fssd, sun2019high, hariharan2015hypercolumns}, feature refinements using non-local attention module \cite{pang2019libra}, gradual multi-stage local information fusions \cite{yu2018deep, sun2019deep}. However, the design of feature fusion is intuitive by directly summing up feature maps after resizing them to the same resolution. Intrinsic properties of the feature pyramid are not explored to let all feature maps contribute equally without distinction. 

Scale-space theory has been studied for decades in traditional computer vision. Effective feature point detection methods \cite{lindeberg1994scale} were proposed by detecting scale-space extrema in the pyramid. Motivated by this, we propose to capture the inter-scale interactions through an explicit convolution in the scale dimension, forming a 3-D convolution in the feature pyramid, termed pyramid convolution (PConv).

\begin{figure}
\begin{center}
\includegraphics[width=0.6\linewidth]{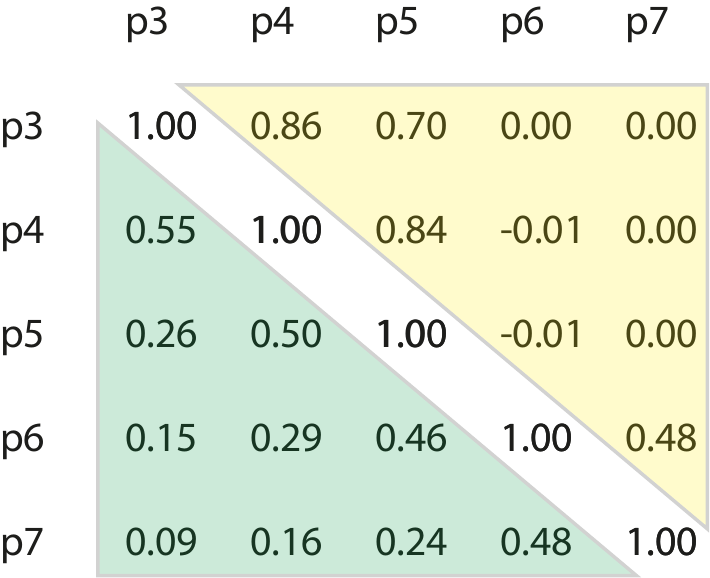}
\end{center}
   \caption{\small{Correlation matrix of feature maps in the feature pyramid of RetinaNet. The upper and lower triangle represent the correlation before and after FPN respectively.}}
\label{fig:correlation}
\end{figure}

Convolution in the scale dimension is a natural choice compare to summing up all feature maps directly. For instance, feature maps of neighboring scales on a feature pyramid should correlate the most, which is however neglected in previous methods. A feature pyramid is built by extracting intermediate outputs after each downsample operation of a feature extraction network (backbone), such as VGG \cite{simonyan2015very}, ResNet \cite{he2016deep} and ResNext \cite{xie2017aggregated}. 
Fig. \ref{fig:correlation} demonstrates the correlation matrix between feature maps extracted from the backbone before and after FPN in RetinaNet. Values close to the diagonal are larger than remote ones. This is similar to the prior of using spatial convolutions for handling natural image that neighboring pixels on an image correlate stronger than distant pairs. However, this property is not directly captured in previous feature fusion designs \cite{pang2019libra, sun2019high}. 

\begin{figure*}[t]
\begin{center}
\includegraphics[width=0.7\linewidth]{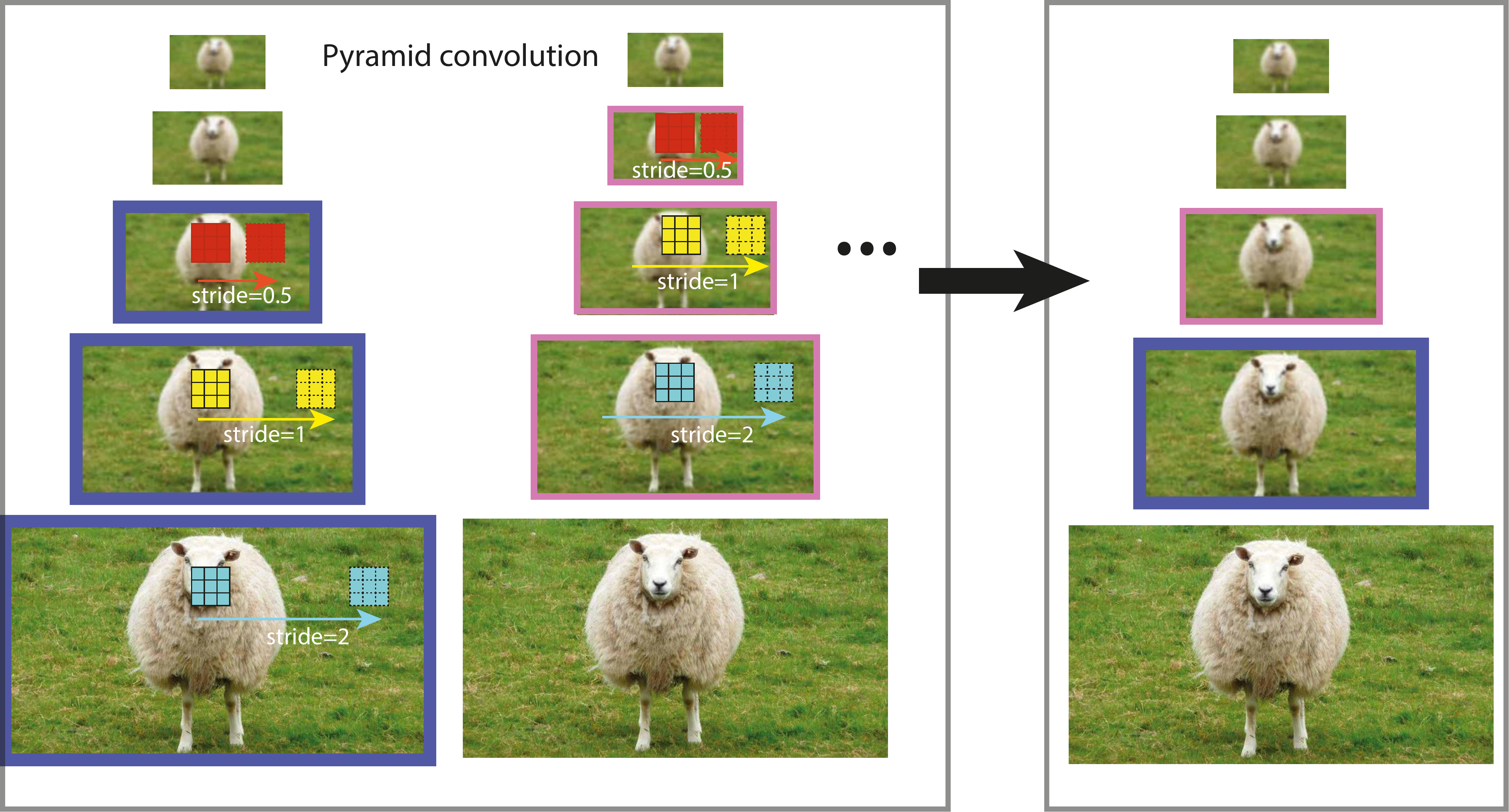}
\end{center}
\caption{\small{Pyramid convolution as a 3-D convolution. Three convolutional kernels (in red, yellow and cyan) are used for this 3-D convolution. The convolutional stride of each kernel scales as the size of the feature map. Feature maps of the same frame color (e.g. blue and pink) generate the feature map at the right side of the same frame color. The image is only used to show the scale and does not represent feature maps.}}
\label{fig:pconv}
\end{figure*}

Further, we also show that the head design of RetinaNet is a special case of PConv with scale kernel = 1, and it actually best suits for extracting features from a Gaussian pyramid. 
A Gaussian pyramid is generated by consecutively blurring an image with a Gaussian kernel followed by a subsampling.
The kernel size of the Gaussian blur should be proportional to the subsampling ratio, so as to remove high-frequency noise during subsampling and not big enough to remove much detail. 
Conducting PConv in this Gaussian pyramid helps extract scale-invariant features.

However, the feature pyramid constructed from a deep backbone network is usually far from a Gaussian pyramid. First, the multiple convolutional layers in the backbone between two feature pyramid levels make a larger effective Gaussian kernel; Second, the theoretical value of effective Gaussian kernel should vary from pixel to pixel due to the non-linearity operations such as ReLU in obtaining the next pyramid features. As a result, we explore the possibility of relaxing these two discrepancies by devising a scale-equalizing module.
Using the idea of deformable convolution \cite{dai2017deformable}, the kernel size at the bottom pyramid is fixed and deforms as the shared kernel strides in the scale dimension. 
Such a modification over PConv now enables it to equalize different pyramid levels (scales) by aligning its kernels when convolving higher layers, and is thus termed as scale-equalizing pyramid convolution (SEPC). 
It can be shown to extract scale-invariance features from feature pyramid and only brings a modest computational cost increase since the deformable kernels are only applied to high-level features. 
Equipped with the SEPC module, the detection performance boosts for various models. For example, SEPC module reaches as high as 4.3AP increase in state-of-the-art single stage detectors, such as FreeAnchor \cite{zhang2019freeanchor}, FSAF \cite{zhu2019feature}, Reppoints \cite{yang2019reppoints} and FCOS \cite{tian2019fcos}, making them even surpass most two-stage detectors. A light version of SEPC (SEPC-lite) can also reach a performance gain of around 3.5AP with only $\sim$7\% increase in computational cost. 

This study mainly contributes in the following aspects. 

(1). We propose a light-weighted pyramid convolution (PConv) to conduct 3-D convolution inside the feature pyramid to cater for inter-scale correlation.

(2). We also develop a scale-equalizing pyramid convolution (SEPC) to relax the discrepancy between the feature pyramid and the Gaussian pyramid by aligning the shared PConv kernel only at high-level feature maps.

(3). The module boosts the performance ($\sim3.5$AP increase on state-of-the-art single stage object detectors) with negligible inference speed compromise.

\section{Related work}
\subsection{Object detection}
Modern object detection architectures are generally divided into one-stage and two-stage ones. Two-stage detection representatives like SPP \cite{he2015spatial}, Fast R-CNN \cite{girshick2015fast}, Faster R-CNN \cite{ren2015faster} first extract region proposals and then classify each of them. The scale variance problem is somewhat mitigated in two-stage detectors where objects of different sizes are rescaled to be the same size during the ROI pooling process. On the other hand, single-stage object detection \cite{liu2016ssd} directly utilizes the intrinsic sliding-window trait of convolutions to build feature pyramids and directly predict objects based on each pixel. Though having earned advantage in real-time tasks due to its fast inference, single-stage detectors has been lagging behind two-stage ones as for the performance. RetinaNet \cite{lin2017focal} is a milestone single-stage detector since it boosts detection performance by adopting focal loss and new design of detection head. Following works further accelerate the model and improve its performance simultaneously by viewing object detection as key point localization tasks and thus removing the dependency on multiple anchors at each feature map \cite{yang2019reppoints, tian2019fcos}. But the design of FPN and head remains the same as RetinaNet. 
\subsection{Feature fusion}
\begin{figure*}[t]
\begin{center}
\includegraphics[width=\linewidth]{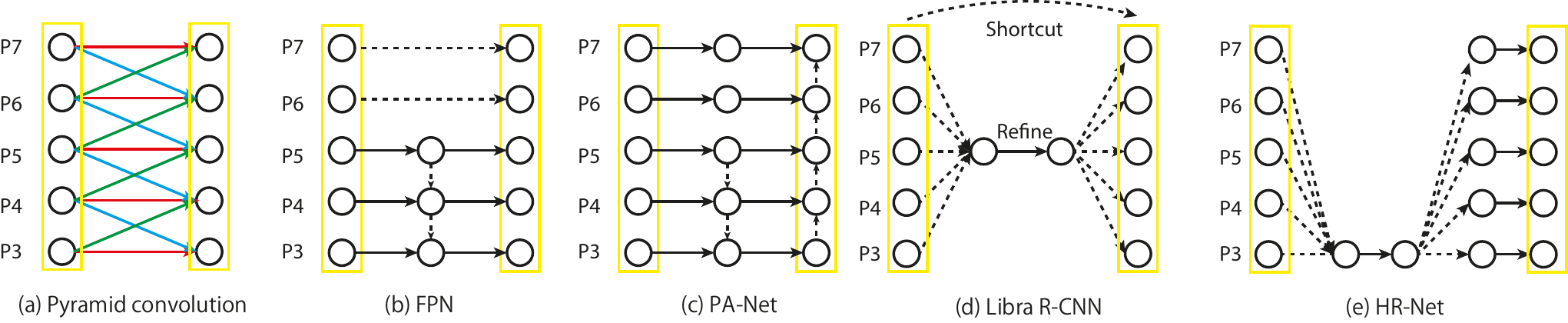}
\end{center}
   \caption{\small{Comparison of pyramid convolution (a) with other feature fusion modules, including (b) FPN \cite{lin2017feature} (c) PA-Net \cite{liu2018path}, (d) Libra R-CNN \cite{pang2019libra} and (e) HR-Net \cite{sun2019high}. Each feature map upward has a spatial sized scaled down by two by default. Dotted lines represent interpolation operations, meaning that they can be upsampling, downsampling or shortcut depending on the respective feature map sizes. Each black solid line is an independent convolution, and colored solid lines of the same color are shared convolution operations.}}
\label{fig:fpn}
\end{figure*}

In deep networks, low-level features are generally deemed lacking in semantic information but rich in keeping geometric details, which is the opposite for high-level features. Therefore, feature fusion plays a crucial rule in combining both semantic and geometric information. Several backbone structures have designs of fusing information from different scales such as Inception network \cite{szegedy2015going} and ScaleNet \cite{li2019data}. FPN \cite{lin2017feature} and its contemporary works leverage high-level feature maps when detecting small objects. The following works further the efficiency of feature fusion from different aspects. As shown in Fig. \ref{fig:fpn}, PA-Net \cite{liu2018path} directly creates a short path for low-level feature maps since detecting large objects also needs the assistance of location-sensitive feature maps. Following the same philosophy, multiple bidirectional information fusion paths were also proposed in \cite{newell2016stacked,zhao2019m2det}. Apart from normal approaches of direct summation, some other methods also adopted concatenation to project all feature maps to a common space followed by a back distribution. Pang et al. \cite{pang2019libra} furthered the level of feature diffusion by adding a non-local block to fine-tune the combined feature maps.  

\subsection{Cross-scale correlation}
There have also been several other methods considering the cross-scale correlation in both traditional and recent research. Cross-scale difference was calculated to approximate the Laplacian operator in SIFT \cite{lowe2004distinctive} to extract scale-invariant features. Worrall \& Welling \cite{worrall2019deep} also extended group convolution to deep neural networks using dilated convolution. Wang et al. \cite{wang2019dsrn} fused feature maps with the neighboring scales to capture inter-scale correlation after all feature maps are transferred to be of the same size as the largest one. In these work, either repeated computations over different transformation of the input image is needed \cite{worrall2019deep} or scale correlation is conducted on a high-resolution feature map \cite{wang2019dsrn}, both of which incur undesirable increase in computational resources. In this study, the pyramidal structure of feature maps is maintained when conducting convolution across different scales, which is much more efficient in computation. Actually, the original design of the head structure of RetinaNet and its descendants can also be viewed as a PConv with scale kernel of 1. Therefore, our design of PConv is compatible with the state-of-the-art single-stage object detectors with minimal computational cost increase. 

\section{Pyramid convolution}
The pyramid convolution (PConv) is indeed a 3-D convolution across both scale and spatial dimensions. If we represent the features in each level as a dot as in Fig. \ref{fig:fpn}a, the PConv can be represented as $N$ different 2-D convolutional kerels. 
Nevertheless, as shown in Fig. \ref{fig:pconv} there is a size mismatch across different pyramid levels. The spatial size is scaled down as the pyramid level goes up. In order to accommodate the mismatch, we set different strides for the $K$ different kernels when convolving in different layers. For example, for PConv with $N=3$, the first kernel should have a stride of $2$ while the last one should have a stride of $0.5$. Then the output of the PConv is 
\begin{equation}
y^l = w_{1} \ast_{s0.5} x^{l+1} + w_{0} \ast x^{l} + w_{-1} \ast_{s2} x^{l-1}, 
\label{eq:pconv}
\end{equation}
where $l$ denotes pyramid level,  $w_1$, $w_0$ and $w_{-1}$ are three independent 2-D convolutional kernels, $x$ is the input feature map and $\ast_{s2}$ means a convolution with stride 2. 
The kernel of stride $0.5$ is further replaced by a normal convolution with stride of 1 and a consecutive bilinear upsampling layer. That is,
\begin{equation}
\label{eq:implement}
y^l = \mathrm{Upsample}(w_{1} \ast x^{l+1}) + w_{0} \ast x^{l} + w_{-1} \ast_{s2} x^{l-1}
\end{equation}
Similar to conventional convolutions, zero-padding is also used for PConv. As for the bottom pyramid level ($l=1$), the last term in Eq. \ref{eq:implement} is unnecessary while for the top-most level ($l=L$), the first term is ignored. Despite the 3 convolution operations at each layer, the total FLOPs of PConv is actually only around 1.5 times as much as the original head (see Appen. 1). 

\subsection{Pipeline}
\begin{figure}
\begin{center}
\includegraphics[width=\linewidth]{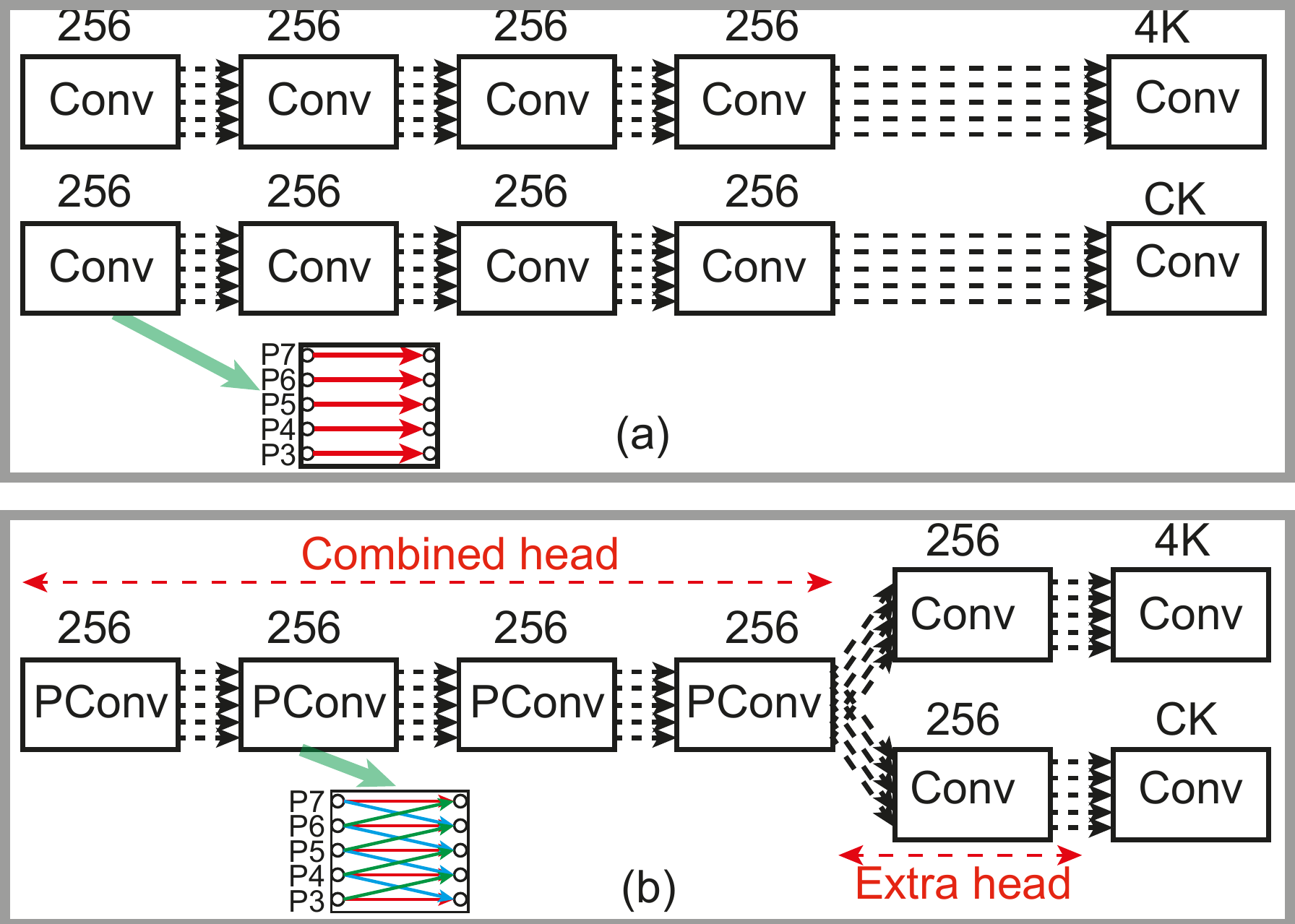}
\end{center}
   \caption{\small{(a) Head design of the original RetinaNet; (b) Head design with PConv. In the final output convolution, $K$ is the number of anchor boxes, which is 1 for anchor-free methods, and $C$ is the number of classes in classification.}}
\label{fig:pipeline}
\end{figure}

Apart from the ability of extracting scale-correlated features, PConv also benefits from its compatibility with the head design of RetinaNet and its descendants. As seen from Fig. \ref{fig:pipeline}a, RetinaNet head is actually also a PConv with a scale kernel of one. Therefore, the 4 convolutional heads can be directly replaced by our PConv module with a scale kernel of 3. The stacked PConv echoes the stacked convolutions modules in 3-D deep networks \cite{tran2015learning}, so as to gradually increase correlation distance without much computational burden. 

However, each PConv still brings some additional computation. As an alternative, the 4 PConv modules are shared by both classification and localization branch, forming a combined head structure as shown in Fig. \ref{fig:pipeline}b. In order to cater for the difference in the classification and localization tasks, an extra normal convolution are also added after the shared 4 PConv modules. It can be calculated that this design has even less FLOPs than the original RetinaNet head (see Appen. 1).
\subsection{Integrated batch normalization (BN) in the head}
\label{subsec:ibn}
In this study, we also retrieve the use of BN in detection head. 
A shared BN follows the PConv module and collects statistics from all feature maps inside the feature pyramid, instead of from a single layer. This design comes naturally as we view PConv as a 3-D convolution. 
Since the statistics are collected from all the feature maps inside the pyramid, the variance becomes smaller, especially for high-level features having small feature map sizes. This enables us to train BN in the head even in a small batch size $\sim 4$ and achieve better performance. 

\section{Scale-equalizing pyramid convolution}
\begin{figure}
\begin{center}
\includegraphics[width=0.8\linewidth]{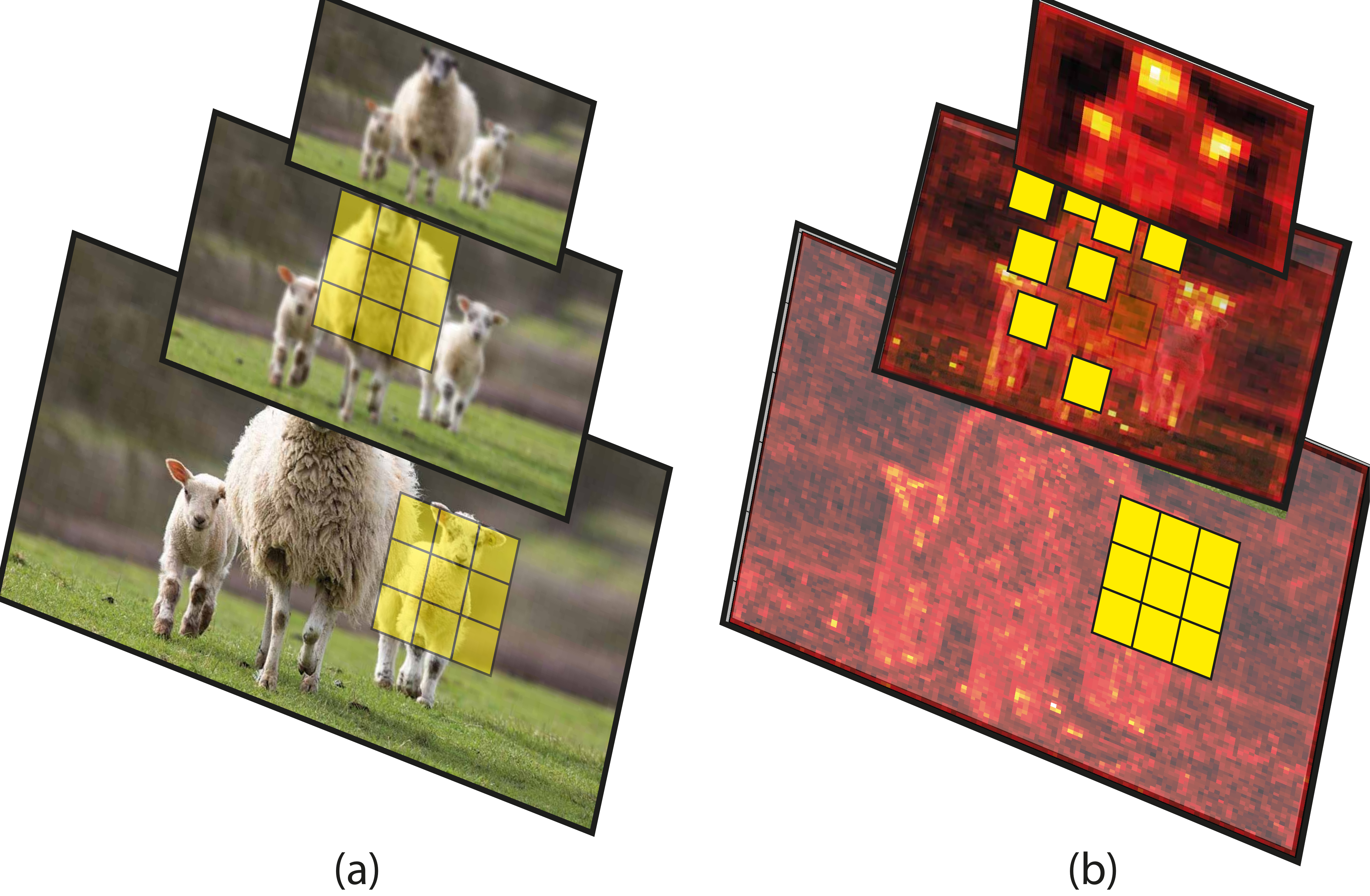}
\end{center}
   \caption{\small{(a) PConv on Gaussian pyramid; (b) SEPC on Feature pyramid}}
\label{fig:sepc}
\end{figure}

In designing the pyramid convolution, we have used a naive implementation. The kernel size of each 2-D convolution used in PConv keeps constant when the kernel strides along the scale dimension, even though the feature map size shrinks. 
This is reasonable when PConv is conducted on a Gaussian pyramid (a Gaussian pyramid is built by consecutively Gaussian blurring an image followed by a downsampling) since 
\begin{remark}
Pyramid convolution is able to extract scale-invariant features from a Gaussian pyramid.
\end{remark}
The detailed mathematical proof can be found in Appen. 3.  It is shown intuitively in Fig. \ref{fig:sepc}a. 
When a PConv with $N=1$ extracts features from the pyramid, objects of different scales can be captured by the same kernel at different level. 
Moreover, the Gaussian blur is also necessary in generating the pyramid so as to avoid high-frequency noises in extracting features in downsampled images. On the other hand, too strong blur conceals details. The optimal blurring kernel in the Gaussian pyramid is around the size of the downsampling ratio between two pyramid levels. 

In the naive implementation of PConv, and also in the design of RetinaNet head, such a fashion is directly used to process feature pyramid. However, the optimal blurring kernel is hardly satisfied for feature pyramid. In Fig. \ref{fig:sepc}b, we see that the blurring effect of feature maps in high-level features becomes much more serious than that in an image pyramid. This is due to the many layers of convolution and non-linearity operations in the backbone between two feature maps in the feature pyramid. 

In order to compromise the stronger blurring effect and extract scale-invariant features, some studies advocated using dilated convolution \cite{worrall2019deep}. That is, as the PConv module strides in the scale dimension, the kernel should also be larger than the one used in the bottom-most features. However, because of the non-linearity operations in the backbone, the dilation ratios of different pixels are also different, making it difficult to directly use a constant one.

Instead, we borrow the idea of deformable convolution to directly predict the offset of the convolutional kernel as the shared kernel strides upward in the scale dimension. As shown in Fig. \ref{fig:sepc}b, the kernel convolving with the the bottom feature map is fixed as a normal $3\times3$ convolution. As it processes high-level feature maps in the feature pyramid, a deformation offset is predicted based on the current layer of feature map. In this way, features in each pyramid level (scale) are equalized by the deformation offset and is ready to be convolved by the shared PConv kernels. Therefore, it is termed scale-equalizing pyramid convolution (SEPC). The pseudo-code for both PConv and SEPC can be found in Appen. 2.

There are multiple benefits in SEPC. 1) The larger blurring effect between two layers of feature pyramid is considered due to its dilating ability of a deformable convolution kernel; 2) The discrepancy of a feature pyramid from the Gaussian pyramid is alleviated.  3) Since the computational cost of a convolution reduces by 4 from one layer to its upper feature pyramid level, adding deformable convolution only to the  high-level feature  maps incurs minimal computations. 
In this study, we study the effect of both SEPC-full that applies SEPC to both the combined and the extra head in Fig. \ref{fig:pipeline}b, and SEPC-lite that applies SEPC only to the extra head.

%
%
%
%
%
%
\section{Experiments}
The experiments in this study were performed on the MS-COCO2017 detection dataset \cite{lin2014microsoft} of 80 categories. The training set is composed of around 118k images, and the validation set consists of 5k images (\texttt{minival}). The detection metrics are reported by default on \texttt{minival}. The results on the test set (\texttt{test-dev}) are also reported for several models in this study. More details in regards to the experimental settings can be found in Appen. 4. 

%

\subsection{Single-stage object detectors}
The single stage object detectors in this study are mostly the latest and state-of-the-art models, using either anchor-based such as RetinaNet, \cite{lin2017focal} and FreeAnchor \cite{ zhang2019freeanchor} or anchor-free methods such as FSAF \cite{zhu2019feature} and Reppoints \cite{yang2019reppoints}.  
The results between the proposed SEPC and their original baselines are compared in Tab. \ref{tab:single_stage}. SEPC-full is found able to boosts the performance by more than 4AP, yet incurring unnecessary inference time increase due to the deformable operations involved. The improvement of SEPC-lite in each network is also substantial, increasing $3.1\sim3.8$AP with only 7\% latency increase. A direct comparison of SEPC-lite on more detectors is shown in Fig. \ref{fig:boosts}. 
It should be noted that our own implemented baseline of FSAF (see details in Appen. 6) already achieves 36.9AP, 1.1AP higher than the original results. And the performance of FSAF is further boosted to 40.9 by SEPC-lite, which is even 0.5AP higher than that of Cascade and Deformable Faster-RCNN while maintaining more than 20\% faster. 
The improvement of SEPC-lite on FreeAnchor, one of the best single-stage detectors, is also surprising, making it achieve 41.7 and painlessly furthering the state of the art by 3.2. 

\subsubsection{Ablation study}
\subsubsection{Effect of each component}
The replacement of normal convolution in the head with PConv brings around $\sim 1.5AP$ increase in various models. As for the speed of PConv, the total FLOPs of PConv is actually smaller than the original head, and the latency still increases by around 3\%, due to the more convolution kernels involved. 

The insertion of integrated BN (iBN) in the head also benefits the model by $0.2\sim 1.2$AP for different architectures. Several other studies also advocated group normalization (GN) when training detection networks \cite{tian2019fcos}. However, one trait of BN that is missing in GN is that BN does not require calculating on-site statistics when conducting inference and can be merged in the former convolutional layer. This brings significant advantage in inference speed, as revealed by the same forward latency with iBN. The increased performance is a natural result of the faster optimization and better generalization of BN. (see Appen. 5)

As for the effect of scale-equalizing module, we also compare the results of SEPC with PConv+iBN and find significant improvement ($1.6\sim 2.5$AP increase), indicating that the scale equalizing module can help align features at different levels and functions well in various object detectors. 

\begin{table*}
\small
\begin{center}
\begin{tabular}{c|c|c|c|c|c|c|c|c|c}
\hline
\hline
Detector  & Note & FLOPS(G) & Time(ms) & AP & AP$_{50}$& AP$_{75}$ & AP$_{S}$&AP$_M$&AP$_L$ \\
\hline
\multirow{6}{*}{RetinaNet} & baseline& 239.32 & 73.2  & 35.7 & 55.0& 38.5&18.9&38.9&46.3 \\
                           & DCN head& \textbf{249.55} & \textbf{102.1} & 36.8 & 56.8& 39.6&20.4&40.3&49.0  \\
                           & PConv& 239.29 & 76.5 & 37.0 & 57.7& 39.4&22.3&40.8&48.9  \\
                           & PConv+iBN& 239.36 & 76.4 & 37.9 & 59.3& 40.6&22.5&42.2&49.1  \\
                           & SEPC& 242.22 & 89.6 & \textbf{39.7} & \textbf{60.4}& \textbf{42.7} &\textbf{23.1}&\textbf{44}&\textbf{52.2}  \\
                           & SEPC-lite & 240 & 78.5 & 38.8 & 59.9& 41.8 &22.6&42.8&51  \\
\hline
\multirow{6}{*}{FSAF} & baseline$^\ast$& 205.2 & 62.4  & 36.9 & 56.1& 39&20.6 & 40.1&48.2 \\
                      & DCN head& \textbf{215.42} & \textbf{85.2}  & 40.1 & 58.5 & 42.8& 22.4& 43.3& 54.7 \\
                      & PConv& 205.18 & 66.0  & 38.7 & 58.9 & 41.1 & 22.2 & 42 & 51 \\
                      & PConv+iBN& 205.25 & 66.1  & 38.9 & 59.1& 41.8&22.2&42.5&51 \\
                      & SEPC& 208.11 & 77.4 & \textbf{41.3} & \textbf{60.4}& \textbf{43.6} & \textbf{23} & \textbf{44.8} & \textbf{57.8} \\
                      & SEPC-lite& 205.88 & 68.2 & 40.7 & 60& 43.4 & 22.4 & 44.6 & 55.1 \\
\hline
\multirow{6}{*}{FreeAnchor}& baseline& 239.32 & 76.4  & 38.5 & 57.3& 41.2 & 21.1 & 41.8 & 51.5 \\
                           & DCN head& \textbf{249.55} & \textbf{100.4}  & 39.4 & 58.0 & 42.4 & 21.7 & 43.0  & 52.7  \\
                           & PConv& 239.29 & 79.4  & 40.0 & 59.1& 43 & 22.8 & 43.8 & 53.3  \\
                           & PConv+iBN& 239.36 & 79.7 & 41.2 & 60.5& 44.3 & 24.3 & 44.6 & 54.6 \\
                           & SEPC& 242.22 & 89.9 & \textbf{42.8} & \textbf{61.9}&\textbf{ 45.9} & \textbf{25.6} & \textbf{46.4} & \textbf{57.4}   \\
                           & SEPC-lite& 240 & 81.2 & 41.7 & 61& 45.1 & 24.2 & 45.2 & 54.8   \\
\hline
\hline
\multicolumn{1}{@{} l}{$^\ast$: our own implementation}
\end{tabular}
\end{center}
\caption{\small{Comparison of detection AP results of different architectures. All models  were trained using ResNet-50 backbone and adopted the \texttt{1x} training strategy. Results were evaluated on COCO \texttt{minival} set.}}
\label{tab:single_stage}
\end{table*}

\begin{figure}
  \centering
  \begin{subfigure}[b]{0.4\linewidth}
	\includegraphics[width=\linewidth]{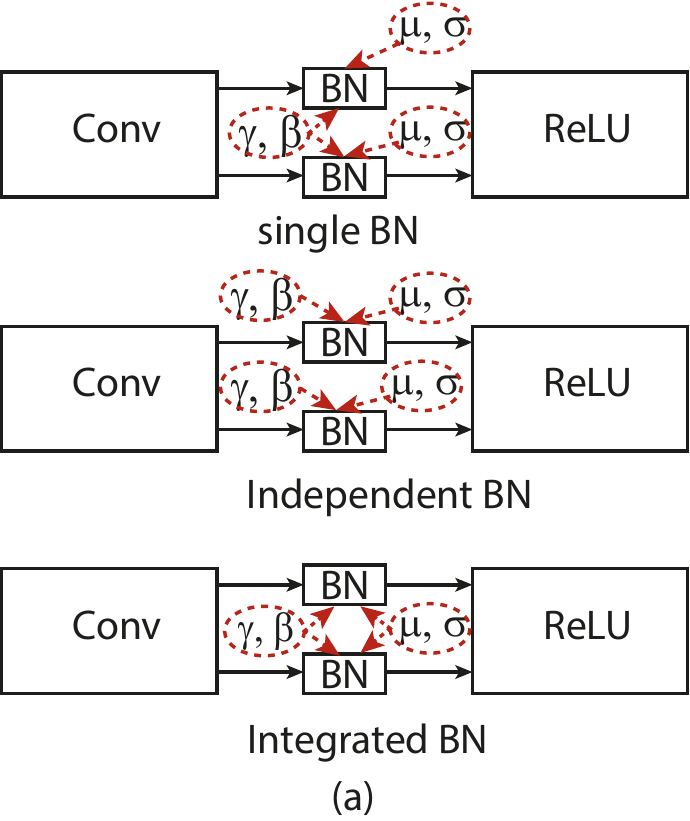}
	\label{fig:bn_mode}
  \end{subfigure}
  \begin{subfigure}[b]{0.57\linewidth}
	\includegraphics[width=\linewidth]{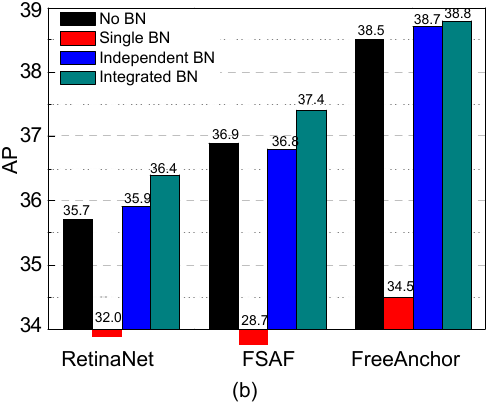}
	\label{fig:bn_effect}
  \end{subfigure}
   \caption{\small{(a) Different batch normalization implementations in the feature pyramid levels. Only 2 feature levels are given as an illustration; (b) Comparison of AP results of architectures with different batch normalization implementations in the feature pyramid.}}
  \label{fig:bn_compare}
\end{figure}

\subsubsection{Comparison with DCN head}
If all convolutions are replaced by deformable convolutional kernels (DCN) \cite{dai2017deformable} in the original head of RetinaNet-alike models (RetinaNet \cite{lin2017focal}, FSAF \cite{zhu2019feature}, FCOS \cite{tian2019fcos}, FreeAnchor \cite{zhang2019freeanchor} et al.), the increase in AP varies. For most models such as RetinaNet and FreeAnchor, the increase in AP is limited ($\sim$ 1AP). The performance increase in FSAF is more significant, possibly due to the combination of the adaptive kernel in DCN and the adaptive loss function designed in FSAF. Nontheless, the runtime costs of DCN head in all these models are huge.  As shown in Tab. \ref{tab:single_stage}, the AP gain of PConv+iBN(no DCN involved) in RetinaNet \& FreeAnchor already significantly outperforms DCN head significantly in both AP and time efficiency. SEPC-lite and SEPC brings further AP gains and outperforms DCN head in all these models, while bringing only $1/5$ and $1/2$ as much runtime overhead.

\subsubsection{Comparison of different BN implementations in the head}
There are different implementations of BN in dealing with feature pyramids, which are shown in Fig. \ref{fig:bn_compare}a. The output after each BN module is $y=\gamma\frac{x-\mu}{\sigma}+\beta$, where $\gamma$  and $\beta$ are parameters, and $\mu$ and $\sigma$ are batch statistics for normalization. 
Single BN adds a BN module after each feature pyramid level with shared parameters $\gamma$ and $\beta$ across the pyramid. But each feature map collects statistics on its own during training. Independent BN makes both parameters and statistics of BN in a feature layer exclusive to itself and is used in \cite{ghiasi2019fpn}. Integrated BN, as discussed in section \ref{subsec:ibn}, calculates batch statistics over all feature maps from feature pyramid networks. 

The effects of these four BN designs are demonstrated in Fig. \ref{fig:bn_compare}. Trivially using single BN results in a catastrophic decline in performance of detectors, due to the mismatch between the shared parameters and non-shared statistics. Integrated BN and independent BN both improves AP, and integrated BN outperforms independent BN because of the more stable statistics during training.


\subsubsection{Comparison with other feature fusion modules}
In regards to different feature fusion methods, Tab. \ref{tab:pconv_fusion} presents our comparison of PConv to other state-of-the-art feature fusion modules on FreeAnchor. It is obvious that PConv provides a dramatic performance increase compared to common feature pyramid networks, including NAS-FPN \cite{ghiasi2019fpn} and Libra \cite{pang2019libra}. Moreover, the designed PConv stack head also earns the minimum FLOPs increase among the feature fusion modules. Results of this section validate the effectiveness of PConv in feature fusion. 	 

\begin{table}
\small
\begin{center}
\begin{tabular}{ c|c|c|c|c }
\hline
\hline
 Feature fusion & AP & AP$_{50}$ &  AP$_{75}$ &FLOPS(G)\\
\hline
 FPN  & 38.5 & 57.3  &41.2 & 239.3\\
 HR-Net  & 38.6 & 57.1 &41.3 & 297.6\\
 PA-Net  & 38.9 & 57.6 &41.6 & 245.9\\
 NAS-FPN & 39.1& 57.0 &41.8 & 347.1\\
 Libra & 39.4 & 58.7  &42.2 & 315.8\\
 PConv & 40.0 & 59.1  &43 &  239.3\\
\hline
\hline
\end{tabular}
\end{center}
\caption{\small{Comparison of PConv with other feature fusion modules including FPN \cite{lin2017feature}, HR-Net \cite{sun2019high}, PA-Net \cite{liu2018path}, NAS-FPN \cite{ghiasi2019fpn} and Libra \cite{pang2019libra} on FreeAnchor. Results evaluated on COCO \texttt{minival} are reported.}}
\label{tab:pconv_fusion}
\end{table}

\subsection{Comparison with state-of-the-art object detectors}
In this section, we compare our method to other state-of-the-art object detectors on COCO2017 benchmarking dataset. The training strategies followed \texttt{2x} with 640-800 scale jitter and the results were obtained with only a single scale, unless specified otherwise.  Details can be found in Appen. 4. We only report FreeAnchor equipped with with SEPC-lite and SEPC for the purpose of real potential applications since SEPC incurs intangible computation cost for large backbones such as ResNext-101. It is observed that SEPC boosts the original baselines by a significant margin and achieves the state-of-the-art 47.7AP using ResNext-101 backbone without utilizing bells and whistles (e.g. multi-scale test, sync BN, deformable backbone), surpassing even most two-stage detectors with deformable backbones and multi-scale test. If DCN backbone and stronger training scale jitter (480-960) is applied, the AP performance reaches 50.1, the best reported detection result on a single-stage model with single-scale test. 

\begin{table*}
\centering
\small
\begin{tabular}{c|c|c|c|cccccc}
\hline
\hline
Method & Backbone  & Epochs & Input Size& AP & AP$_{50}$ & AP$_{75}$& AP$_s$ & AP$_m$ & AP$_l$\\
\hline
\multicolumn{10}{@{} l}{Two-Stage Detectors}\\
\hline
Cascade-RCNN$^-$ \cite{cai2018cascade} & ResNet-101  & 18 & min800 & 42.8 & 62.1 & 46.3 & 23.7 & 45.5 & 55.2 \\
TridentDet \cite{li2019scale} & ResNet-101 & 24 & min800 & 42.7 & 63.6 & 46.5 & 23.9 & 46.6& 56.6 \\
SNIP$^\ast$ \cite{singh2018analysis} & DCN+ResNet-101 & - & - & 44.4 & 66.2 & 49.9 & 27.3 & 46.4 & 56.9 \\
SNIPPER$^\ast$ \cite{singh2018sniper} & DCN+ResNet-101 & - & - & 46.1 & 67.0 & 51.6 & 29.6& 48.9& 58.1 \\
TridentDet$^{\square\triangle}$ \cite{li2019scale} & DCN+ResNet-101 & 36 & min800 & 46.8 & 67.6 & 51.5 & 28.0 & 51.2& 60.5 \\
\hline 
\multicolumn{10}{@{} l}{One-Stage Detectors}\\
\hline 
FreeAnchor \cite{zhang2019freeanchor} & ResNet-101 & 24 & min800 & 43.0 & 62.2 & 46.4 & 24.7 & 46& 54 \\
FSAF \cite{zhu2019feature} & ResNext-101-64x4d & 24 & min800 & 44.1 & 64.3 & 47.3 & 26.0 & 47.1 & 55.0 \\
FreeAnchor \cite{zhang2019freeanchor} & ResNext-101-64x4d & 24 & min800 & 44.9 & 64.4 & 48.4 & 26.5 & 48 & 56.5 \\
AlignDet \cite{chen2019revisiting} & ResNext-101-32x8d & 18 & min800 & 44.1 & 64.7 & 48.9 & 26.9 & 47.0 & 54.7 \\
CornerNet \cite{law2018cornernet} & Hourglass-104 & $\sim$200 & 511 & 40.6 & 56.4 & 43.2 & 19.1 & 42.8 & 54.3  \\
CenterNet \cite{duan2019centernet} & Hourglass-104 & $\sim$200 & 511 & 44.9 & 62.4 & 48.1 &25.6 & 47.4 & 57.4 \\
NAS-FPN \cite{ghiasi2019fpn} & AmoebaNet + Drop Block & 150 & 1280 & 48.3 & - & - & - & - & -  \\
\hline

\textbf{FreeAnchor+SEPC-lite} & ResNet-101 & 24 & min800 & \textbf{45.5}& 64.9 & 49.5 & 27 & 48.8 &56.7  \\
\textbf{FreeAnchor+SEPC-lite} & ResNext-101-64x4d & 24 &min800 & \textbf{47.1} & 67.0 & 51.2 & 29.3 & 50.8 & 58.3   \\
\textbf{FreeAnchor+SEPC} & ResNext-101-64x4d & 24 &min800 & \textbf{47.7} & 67.3 & 51.7 & 29.2 & 50.8 & 60.3   \\
\textbf{FreeAnchor+SEPC} $^\dagger$ & DCN+ResNext-101-64x4d & 24 &min800 & \textbf{50.1} & 69.8 & 54.3 & 31.3 & 53.3 & 63.7   \\
\hline
\hline
\multicolumn{10}{@{} l}{$^\ast$: Multi-scale testing; \quad $^-$: single-scale training; \quad $^\square$: soft-nms;\quad  $^\triangle$: synchronized BN} \quad $^\dagger$: wider training scales(480-960)
\end{tabular}
\caption{\small{Comparing of the single-model \& single-scale test results of SEPC with other state-of-the-art object detectors. Results are evaluated on \texttt{test-dev}}. 
}
\label{tab:compare}
\end{table*}

\subsection{Extension to two-stage object detectors}
We also present that PConv (without scale-equalizing module) can still be effective when it is applied to two-stage object detectors. As shown in Tab \ref{tab:two_stage}, PConv provides remarkable improvement of AP on different two-stage detectors. PConv provides the most AP increase to Mask-RCNN, which improves the AP by 2.3. 								 

\begin{table}
\small
\begin{center}
\begin{tabular}{c|c|c|c|c}
\hline
\hline
Detector & Note  & AP & AP$_{50}$ & AP$_{75}$\\
\hline
\multirow{2}{*}{Faster} & Baseline  & 36.5 & 58.4 & 39.1 \\
                        & PConv  & 38.5 & 59.9 & 41.4 \\
\hline
\multirow{2}{*}{Mask} & Baseline & 37.3 & 59 &40.2\\
                      & PConv & 39.6 & 60.1 & 43.5 \\
\hline
\multirow{2}{*}{HTC} & Baseline& 42.1 & 60.8 & 45.9 \\
                     & PConv& 43.6 & 62.0 & 47.4 \\
\hline
\hline
\end{tabular}
\end{center}
\caption{\small{Extension of only PConv module to two-stage detectors including Faster R-CNN \cite{ren2015faster}, Mask R-CNN \cite{he2017mask} and HTC \cite{chen2019hybrid}.}}
\label{tab:two_stage}
\end{table}

\section{Conclusion}
In this study, we explore considering the inter-scale correlation through a pyramid convolution (PConv), which runs a 3-D convolution on both the scale and spatial dimension of the feature pyramid. The striding pattern for this PConv in both spatial and scale dimension is quite different from conventional ones. First, due to the different spatial sizes in the pyramid, the striding step of the spatial slices of PConv kernels is proportional to the convolved feature map size in the pyramid level. This pendulum-alike striding pattern of the PConv kernel helps align the spatial position of neighboring feature maps as they are involved in one PConv. Second, when PConv strides up in the scale dimension, the kernel should also adjust its spatial deformation as well, which is then called scale-equalizing pyramid convolution (SEPC).  The naive striding pattern with a fixed spatial kernel size actually best suits for extracting features in a Gaussian pyramid, which is quite far from the feature pyramid generated by deep networks. And SEPC helps relax this discrepancy and extracts more robust features. Being light-weighted and compatible with most object detectors, SEPC is able to significantly improve the detection performance with minimal computational cost increase. 
\section{Acknowledgement}
We would like to thank Wenqi Shao and Zhanghui Kuang from SenseTime for the inspiring discussions and suggestions.
\newpage
{\small
\bibliography{egbib}
}

\end{document}


\title{\textit{Supplementary material for}\\ Scale-equalizing Pyramid Convolution for object detection}

\author{Xinjiang Wang\thanks{equal contribution}, Shilong Zhang$^\ast$, Zhuoran Yu, Litong Feng, Wayne Zhang\\
Sensetime Research\\
{\tt\small \{wangxinjiang, zhangshilong, yuzhuoran, fenglitong, wayne.zhang\}@sensetime.com}
}

\maketitle

\begin{abstract}
We present details of FLOPS calculation in Sec. \ref{sec:flops}. The pseudo-code of pyramid convolution is in Sec. \ref{sec:code}. The detailed mathematical proof that pyramid convolution can extract scale-invariant features from Gaussian pyramid is in Sec. \ref{sec:derivation}. Sec. \ref{sec:details} shows the details of experiments including ablation study and test of inference time respectively.  Extra ablation experiments of integrated batch normalization(iBN) and the effect of the number of PConv layers are presented in Sec. \ref{sec:ablation}. Finally, We present implementation details of Feature Selective Anchor Free module in Sec. \ref{sec:fsaf}. 
\end{abstract}

\section{FLOPs in the head}
\label{sec:flops}
\begin{figure}[t]
\begin{center}
\includegraphics[width=0.7\linewidth]{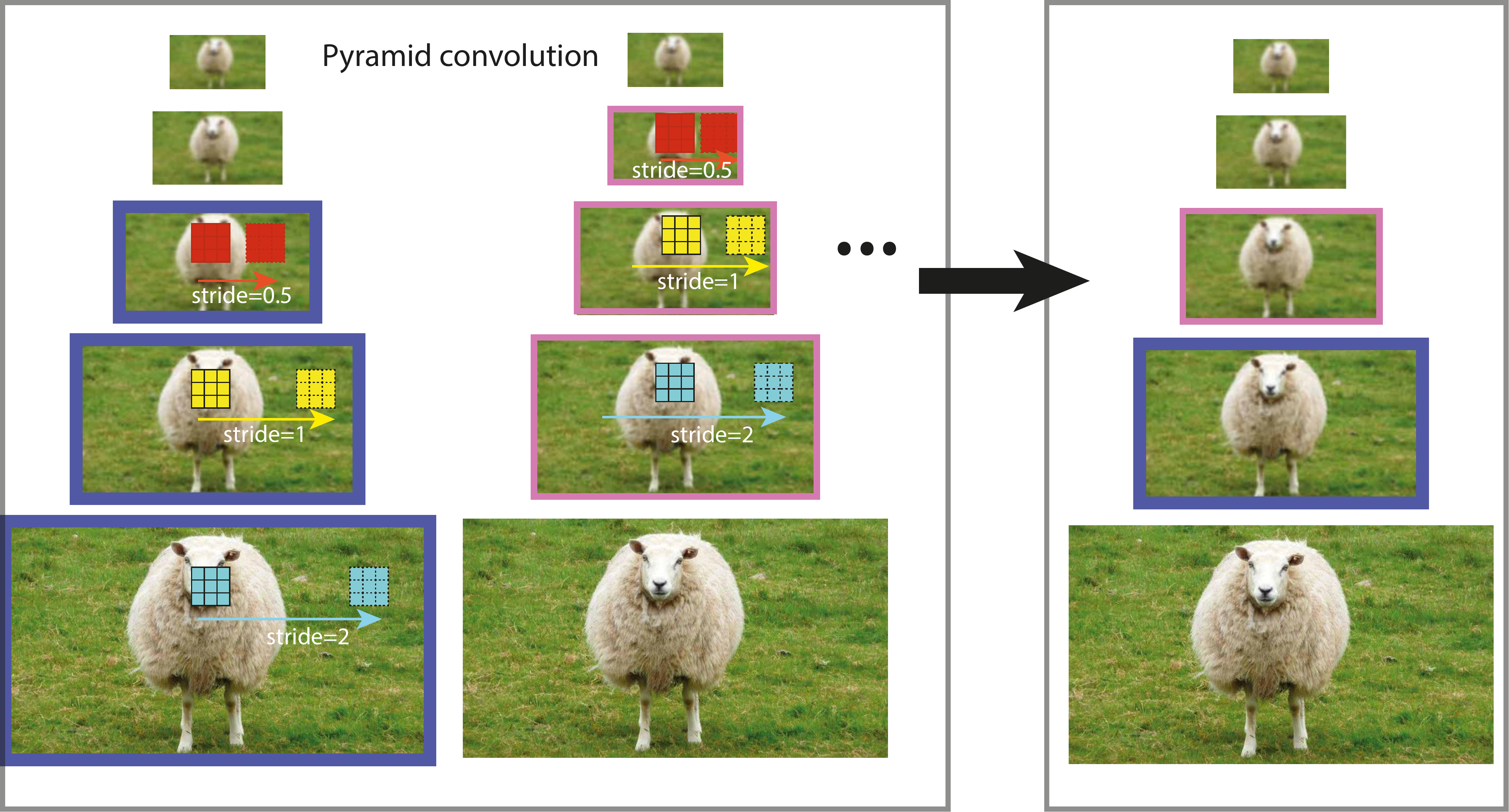}
\end{center}
\caption{Illustration of pyramid convolution on the feature pyramid}
\label{fig:pconv}
\end{figure}

The input image size of all models is $3 \times 1280 \times 800$. we follow the computation of FLOPs in mmdetection\cite{mmdetection} for normal convolution, where one multiplication-addition pair is counted for a single operation.  This yields 
\begin{equation}
\mathrm{FLOPs}=C_{in} \times K_h \times K_w \times H \times W \times C_{out}
\end{equation}
for a single convolution operation, where $C_{in}$ and $C_{out}$ are the number of input and output channels of a convolution, which are all fixed as 256 in this study, $K_h$ and $K_w$ are the convolutional kernel sizes and set as 3, $H$ and $W$ are the width and height of a feature map.

PConv fuses features of adjacent levels to the work-on level by 3-D convolutions with stride of 0.5(for upper level) and of 2(for lower level). For the convolution with stride of 0.5, we implemented by performing a regular convolution with stride of 1 followed by an upsample operation, which can be represented in formula as:
\begin{equation}
\label{eq:pconv}
y^l = \mathrm{Upsample}(w_{1} * x^{l+1}) + w_0*x^l + w_{-1} *_{s2}x^{l-1},
\end{equation}
where $x^l$ is the feature map at level $l$ and $w_1, w_0, w_{-1}$ are three independent convolutional kernels. Since the feature map size of $x^{l+1}$ is half of $x^{l}$ and the size of $x^{l-1}$ is twice as much as $x^{l}$, the computational cost of the first term in Eqn.\ref{eq:pconv} is a quarter of that of the second term whereas the last two terms share the same computational cost.  Note that at top-most level(P7), the first term is eliminated and at bottom-most level(P3),  the last term is eliminated.  We abbreviate the analysis of upsample operation as its cost is relatively small compared to convolutions in this place.  Therefore, the ratio $c_i$ of the FLOPs after applying PConv to that of the original ones in feature pyramids are 2, 2.25, 2.25, 2.25, 1.25 in top-down order. Additionally, the FLOPS associated with each level is also proportional to $H \times W$. The ratio between the spatial size of each feature map and total spatial size can be represented by $ r_i = \frac{H_i \times W_i}{\sum_{j=3}^{7} H_j \times W_j}$. The numeric values of such ratios are calculated as: 0.0029,0.0117,0.0469,0.1877,0.7507
in top-down order. Thus, the total computation of using 4 stacked PConv is $ C = \sum_{i=3}^{7} c_i \times r_i = 1.4985$ times of using 4 stakced convolution in regular settings.

When classification and regression subsets use combined PConv structure with one extra non-combined for each subnet, the total computation is $\frac{(4 \times C + 2 \times 1)}{2\times4\times1} = \mathbf{0.99925}$ times of that using default head design. 

When SEPC or SEPC-Lite is used, we discuss how FLOPs in deformable conv are calculated. One forwarding of deformable conv is composed of a normal convolution with output channels of $2 \times K_h \times K_w$ for offset prediction, a bilinear interpolation for every sampling point which involves 8 matiplications and 7 additions, and another normal convolution. So we get 
\small
$$FLOPS  \approx  (1 + \frac{ 8 + 2 \times K_h \times K_w }  {C_{out}})\times C_{in} \times K_h \times K_w \times H \times W \times C_{out}$$

In SEPC-Lite, we use a normal conv2D on P3 and deform it through P4-P7. Notice that $\frac{\sum_{i=4}^{7} H_i \times W_i}{\sum_{i=3}^{7} H_i \times W_i} \approx 0.249$, using each deformable convolution only introduces  $0.249 \times \frac{26}{256} = 0.025$ times more computation comparted to using normal conv in feature pyramid. When it comes to SEPC, the actual method for evaluating computation cost is slightly different from calculation in SEPC-Lite due to P3 and P7, however, following a similar trajectory, one can easily justify that the extra computation cost is still marginal. 

\section{Implementation pseudocode of PConv and SEPC}
\label{sec:code}
The pseudocodes of pyramid conv (PConv) are attached as follows, which also corresponds to Fig. 3 \& 4(a) in the manuscript. Note that only normal Conv2D modules are used in pconv.

\begin{lstlisting}[language=Python]
def pconv_module_forward(x, conv2D_list):
	# x: input feature list [p3,p4,p5,p6,p7]
	# conv2D_list: conv2D module list, 
	# [nn.Conv2D(stride=2),nn.Conv2D(),nn.Conv2D()]
	out_x = []
	for level in range(len(x)):
		tmp = conv2D_list[1](x[level]) 
		if level > 0:
			tmp += conv2D_list[0](x[level-1]) 
		if level < len(x) - 1:
			tmp += Upsample(conv2D_list[2](x[level+1])                                                
		out_x.append(tmp)
	return out_x
\end{lstlisting}

As for Scale-equalizing pyramid conv (SEPC), within the \lstinline{for} loop of the pseudocodes, \lstinline{conv2d_list[i]} changes to \lstinline{DeformableConv(conv2d_list[i].weight)}, where \lstinline{i} $\in \{0,1,2\}$, \textbf{only when} \lstinline{level > 0} \lstinline{} (i.e. excluding P3 layer). The main idea is illustrated in Fig. 6 in the manuscript. Note that when convolving the lowest-level features (P3), a normal Conv2D is utilized, whose weights are shared by deformable convs used in higher layers. 

Therefore, SEPC is an improved version of pconv, to relax the discrepancy of feature pyramid from a Gaussian pyramid by aligning the feature map of higher layers with the lowest layer.
In one word, pconv only uses plain Conv2D modules while SEPC leverages deformable convs in an efficient way. Different from a naive implementation of deformable convolution in the head, SEPC only deforms the kernel when convolving higher layers, which is motivated by the perspective from scale space theory and is much more effective w.r.t. computational cost gain.

\section{Discussion about remark 1}
\label{sec:derivation}
\begin{remark}
Pyramid convolution is able to extract scale-invariant features from a Gaussian pyramid.
\end{remark}

\textbf{Gaussian scale space.} Consider an image $f: \mathbb{Z}^2 \rightarrow \mathbb{R}^2$, where the input domain represents the pixel coordinate and $f(\mathbf{x})$ is pixel intensity, a Gaussian scale space
is generated by consecutively blurring the initial image $f_0$ with an isotropic 2-D Gauss-Weierstrass kernel $G(\mathbf{x},t)=(4\pi t)^{-1}\mathrm{exp}({\left\|\mathbf{x}\right\|^2/4t})$ of variable width $\sqrt{t}$ and spatial position $\mathbf{x}$.  A set of responses $f(t,x),\,t\geq 0$ represents blurred images, forming a Gaussian scale space \cite{lindeberg1994scale} (GSS), as written by:
\begin{equation}
    f(t,\textbf{x})=[G(\cdot,t)\ast f_0](\mathbf{x}), \quad t>0
\end{equation}
where a higher $t$ indicates a larger blur. 

\textbf{Gaussian pyramid.} With the above introduction of GSS, a Gaussian pyramid is denoted as 
\begin{equation}
p(a, \mathbf{x}) = f(t(a, s_0), a^{-1} \mathbf{x})
\end{equation}
where $s_0$ is the initial scale, $a$ is the downsizing ratio $0<a\leq 1$ and 
\begin{equation}
\label{eq:scale}
t = \frac{s_0}{a^2} - s_0
\end{equation}
is the Gaussian kernel variance corresponding to the downsizing ratio $a$ in order to keep the same frequency limit after downsizing\cite{worrall2019deep}.  
In practice, $a$ is chosen to be $2^{-l}$, where $l$ is the level of Gaussian pyramid with $l=0$ denoting no sub-sampling on the original image. Then the Gaussian pyramid is also written as
\begin{equation}
p_l(\mathbf{x}) = f(t(2^{-l}, s_0), 2^{l} \mathbf{x}).
\end{equation}


In fact, we can also define an action $S_n$ that transfer from the original level into another by,
\begin{equation} 
\label{eq:action}
[S_{n}[p]](\mathbf{x}) = p_{n}(\mathbf{x}) = [G(\cdot, t(2^{-n}, s_0))\ast p_0](2^{n}\mathbf{x}).
\end{equation}
\begin{lemma}
\label{lemma:associativity}
The actions $S_m$ and $S_n$ satisfy $S_{m} S_{n}=S_{m+n}$
\end{lemma}

\begin{proof}
Since the sub-sampling ratio $2^m2^n=2^{m+n}$ naturally satisfies this argument, we mainly focus on the Gaussian convolution in Eqn. \ref{eq:action}. In order to prove the associativity property of $S_n$ and $S_m$, we only need to calculate the associativity property of convolution 
\begin{equation}
\label{eq:combine}
(G(\cdot, t(2^{-m}, s_0))\ast [G(\cdot, t(2^{-n}, s_0))\ast f(\cdot)](2^n\mathbf{x}))(2^m\mathbf{x}) = [G(\cdot, t(2^{-m-n}, s_0))\ast f(\cdot)](2^{m+n}\mathbf{x})
\end{equation}
This process contains four sub-processes: (1) Gaussian convolution with variance $t(2^{-n})$, (2) sub-sampling by $2^n$; (3) Gaussian convolution with variance $t(2^{-m})$; (4) sub-sampling by $2^m$.
Then we will follow this process and apply Fourier transform on after another.
Applying Fourier transform to $G(\cdot, t(2^{-n}, s_0))\ast f(\cdot)$, we obtain
\begin{equation}
\mathcal{F}_1 = \mathcal{F}\left([G(\cdot, t(2^{-n}, s_0)\ast f(\cdot)])\right) = \exp(-t(2^{-n}, s_0)\left\|\mathbf{\omega}\right\|^2))\mathcal{F}_0(\omega),
\end{equation}
where $\mathbf{\omega}\in\mathbb{R}^2$ represents 2-D frequency in the Fourier-transformed domain, $\mathcal{F}_0(\omega)$ is the Fourier transform of $f(\mathbf{x})$.
Substituting Eqn. \ref{eq:scale} into the above equation acquires 
\begin{equation}
\mathcal{F}_1=\exp((2^{2n} s_0 - s_0)\left\|\mathbf{\omega}\right\|^2)) \mathcal{F}_0(\omega),
\end{equation}

After process (2) (sub-sampling by $2^n$), the Fourier transform is now 
\begin{equation}
\mathcal{F}_2 =\mathcal{F}\left([G(\cdot, t(2^{-n}, s_0)\ast f(\cdot)])(2^n\mathbf{x})\right) = 2^{-n} \exp(-t(2^{-n}, s_0)\left\|2^{-n}\mathbf{\omega}\right\|^2))\mathcal{F}_0(2^{-n}\omega) =2^{-n}\exp((s_0 - 2^{-2n}s_0)\left\|\mathbf{\omega}\right\|^2))\mathcal{F}_0(2^{-n}\omega),
\end{equation}
As for process (3) (Gaussian convolution with variance $t(2^{-m})$), 
\begin{equation}
\mathcal{F}_3 = \mathcal{F}\left([G(\cdot, t(2^{-m}, s_0)])\right)\cdot \mathcal{F}_2 = 2^{-n} \exp((2^{2m} s_0 - 2^{-2n} s_0)\left\|\mathbf{\omega}\right\|^2))\mathcal{F}_0(2^{-n}\omega),
\end{equation}
As for process (4) (sub-sampling by $2^{-m}$), 
\begin{equation}
\mathcal{F}_4 = 2^{-m-n} \exp((s_0 - 2^{-2n-2m} s_0)\left\|\mathbf{\omega}\right\|^2))\mathcal{F}_0(2^{-m-n}\omega),
\end{equation}

It is obvious that after process 4), the Fourier-transformed expression is equivalent to the Fourier transform of RHS of Eqn. \ref{eq:combine}
\begin{equation}
\mathcal{F}\left([G(\cdot, t(2^{-m-n}, s_0)\ast f(\cdot)])(2^{m+n}\mathbf{x})\right) = 2^{-m-n} \exp((s_0 - 2^{-2n-2m} s_0)\left\|\mathbf{\omega}\right\|^2))\mathcal{F}_0(2^{-m-n}\omega) = \mathcal{F}_4
\end{equation}
\end{proof}

The above lemma is very useful in a Gaussian pyramid, since it means that one level in the Gaussian pyramid is able to be transferred to another by a simple \textit{jumping} action, such that 
\begin{equation}
\label{eq:transfer}
p_{m+n} = S_m[p_n]
\end{equation}
Now recall that the expression of pyramid convolution is given by
\small
\begin{equation}\label{eq:pyraconv}
    y^l=w_{1}\ast_{s0.5} x^{l+1}+w_{0}\ast x^{l}+w_{-1}\ast_{s2} x^{l-1},
\end{equation}
where $w_1$, $w_0$ and $w_{-1}$ are three independent kernels, $\ast_2$ denotes a convolution with stride 2, $x^l$ represents the feature pyramid in the $l^{th}$ layer. Once the feature pyramid $x^l$ can be viewed as a Gaussian pyramid, the pyramid convolution is written as 

\begin{equation}
y^{l}(\mathbf{z}) = \sum_{k=-1}^1 [w_{k}\ast p_{l+k}](2^{-k}\mathbf{z}) =\sum_{k=-1}^1\sum_{\mathbf{u}\in\mathbb{Z}^d}w_{k}(\mathbf{u})p_{l+k}(\mathbf{u}+2^{-k}\mathbf{z}).
\end{equation}
Note that the stride option at neighboring layers is now represented by $2^{-k}$. For $k=-1$ with larger feature map size, the stride is 2 and for $k=1$ with smaller feature map size, the stride is now $\frac{1}{2}$.
If we apply a \textit{jumping} action $S_m$ on the output PConv and leverage Eq. \ref{eq:transfer},
\begin{equation}
\begin{split}
S_m(y^l(\mathbf{z})) &= S_{m}\left[\sum_{k=-1}^1 [w_{k}\ast [p_{l+k}]](2^{-k}\mathbf{z})\right]  \\
&=\sum_{k=-1}^1 [w_{k}\ast [p_{l+k+m}]](2^{m-k}\mathbf{z}) \\
&= \sum_{k=-1}^1 w_{k}\ast [S_{m}[p_{l+k}]](2^{-k}\mathbf{z}) \\
\end{split}
\end{equation}

The above equation shows an important property of using PConv in a Gaussian pyramid. That is, PConv commutes with the \textit{jumping} action on the pyramid, which is conventionally called scale equivariance. It can be rephrased in another way. When the scale of an object changes in the original image, the extracted feature can also be found by shifting the convolved pyramid after using PConv, which also fits into the usual definition of scale invariance in object detection\cite{lin2017feature}. 
\section{Experiment details}
\label{sec:details}
\subsection{Training details}
We trained the model with backbone ResNet-50 and ResNet-101 and mini-batch size of 16 on 8 Nvidia Titan XP GPUs. The training budget for the strategy \texttt{1x} was 12 epochs. The initial learning rate was 0.01, and was decreased by 0.1 after 8 and 11 epochs. When the \texttt{2x} schedule was adapted, we used 24 epochs for training and kept the same learning rate and decreased it by 0.1 after 16 and 22 epochs.  All models with ResNext101-64-4d backbone were trained on Nvidia V100 GPUs under the same setting. When using BN in all experiments,we set 4 images per gpu with the same batch size to get more accurate statistics.

In the experiments of evaluating other feature fusion modules, all models used the same backbone ResNet-50 with 1x schedule. We used 4 PConvs in a combined way with one extra head to get better trade-off.In the HRNet,PANet, and Libra, and only replaced the origin FPN with the feature fusion module in the origin paper. In NAS-FPN,we used 7 merging-cells and keep channel 256.

\subsection{Speed test details}
 We compared the speed of  our method (include pre-precossing, forwarding and nms) with other proposed one stage detectors. All evaluation was performed on one Nvidia 1080Ti GPU with  i7-7700k@4.2GHz.  We set batch size to 8 and started the timer at 100-th iteration to make sure I/O  is stable. Then we used the means  of next 200 iteration in computation of speed.
 
\section{Supplementary ablation experiments}
\label{sec:ablation}
\subsection{Effect of the number of pyramid convolution stacks}
\begin{figure}
\begin{center}
\includegraphics[width=0.7\linewidth]{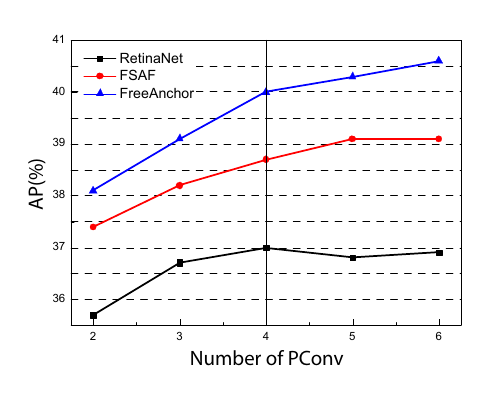}
\end{center}
   \caption{AP change with number of pyramid convolution in different architectures.}
\label{fig:npconv}
\end{figure}

The total number of pyramid convolutions is adjusted from 2 to 6. The recorded AP of different detectors is shown in Fig. \ref{fig:npconv}. The figure illustrates that all these three detectors benefit from increasing the number of PConv from 2 to 4 due to the gradual information flow from top to bottom by PConv stacks.  Using four stacked PConv in heads is rational as it provides descent average precision without causing much redundancy. 

\subsection{Training curves using iBN}
\begin{figure}[b]
\begin{center}
\includegraphics[width=\linewidth]{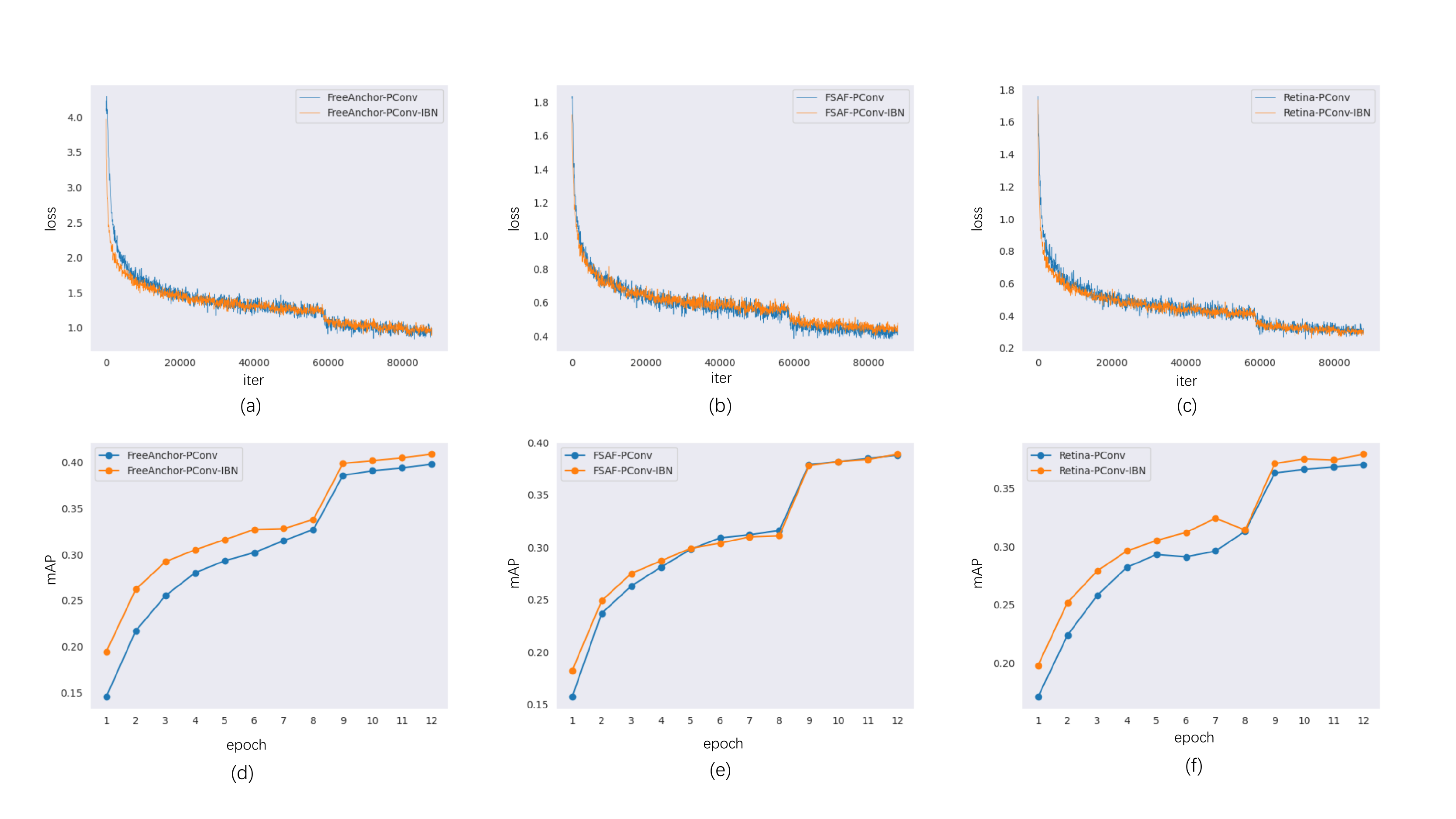}
\end{center}
\caption{Comparison of training loss and validation AP with and without iBN. (a-c) shows the training loss of FreeAnchor, FSAF and RetinaNet, and (d-f) are the validation AP of FreeAnchor, FSAF and RetinaNet, respectively.}
\label{fig:ibn}
\end{figure}

Fig. \ref{fig:ibn}(a-c) displays the training loss of different models and Fig. \ref{fig:ibn}(d-f) presents how AP changes as training goes. 

When iBN is used, in general the losses reduce faster in the early stage of training especially for FreeaAchor and RetinaNet. However, at the end of training, models with iBN result in a slightly higher training loss, yet the mAP is higher than models without IBN. This observation follows the better generalization property of batch normalization.

\section{Details of FSAF}
\label{sec:fsaf}
We only implemented the anchor-free branch in original paper.  In the re-implementation process, we found that in label assigning phase,  removing the ignore region could effectively improve model's ability to distinguish between positive and hard negative samples. Thus, in experiments,  we set effective area as regions within 0.2 from center of projected area of object on the feature map, and assigned negative labels to all outside areas. This effort resulted in 1.1mAP improvement.
%

{\small
\bibliographystyle{ieee_fullname}
\bibliography{egbib}
}